**Article Title**

Benchmark dataset and instance generator for Real-World Three-Dimensional Bin Packing Problems.

**Authors**

Eneko Osaba[1], Esther Villar-Rodriguez[1] and Sebastián V. Romero[1]

**Affiliations**

1. TECNALIA, Basque Research and Technology Alliance (BRTA), 48160 Derio, Spain

**Corresponding author(s)**

Eneko Osaba (eneko.osaba@tecnalia.com)

**Abstract**

In this article, a benchmark for real-world bin packing problems is proposed. This dataset consists of 12 instances of varying levels of complexity regarding size (with the number of packages ranging from 38 to 53) and user-defined requirements. In fact, several real-world-oriented restrictions were taken into account to build these instances: *i)* item and bin dimensions, *ii)* weight restrictions, *iii)* affinities among package categories *iv)* preferences for package ordering and *v)* load balancing. Besides the data, we also offer an own developed Python script for the dataset generation, coined *Q4RealBPP-DataGen*. The benchmark was initially proposed to evaluate the performance of quantum solvers. Therefore, the characteristics of this set of instances were designed according to the current limitations of quantum devices. Additionally, the dataset generator is included to allow the construction of general-purpose benchmarks. The data introduced in this article provides a baseline that will encourage quantum computing researchers to work on real-world bin packing problems.

**Specifications Table**

| **Subject** | Artificial Intelligence |
|---|---|
| **Specific subject area** | Quantum Computing, Discrete Mathematics and Optimization |
| **Type of data** | Text Files, Python File, Figures, Tables |

| How data were acquired | The whole benchmark has been generated using *Q4RealBPP-DataGen*, an automatic instance generator developed ad hoc for this research. The generator, implemented in Python 3.9, automatically saves the instance files in *.txt* format. The packages that compose each solution are generated following the size distribution proposed in [1]. |
|---|---|
| **Data format** | Raw, Analyzed |
| **Description of data collection** | The data has been generated in a laboratory environment through the *Q4RealBPP-DataGen* script. The dataset is useful to validate solvers against industrial use cases: item sizes are compliant with the proposal presented in [1], and a list of real-world oriented requirements is specified (activated or deactivated) for further analysis on problem complexity. Furthermore, the *Q4RealBPP-DataGen* data generator is also provided, which allows the user to create new instances to enrich the evaluation with customized use cases. |
| **Data source location** | The data has been synthetically generated by means of the generator in a laboratory located in TECNALIA, Basque Research and Technology Alliance (BRTA), 48160 Derio, Bizkaia, Spain. The information contained in the benchmark instances has no geographic reference. |
| **Data accessibility** | The whole dataset and the *Q4RealBPP-DataGen* generator are available in a Mendeley public repository: *Repository Name*: Benchmark dataset and instance generator for Real-World 3dBPP. *Data identification number:* doi: 10.17632/y258s6d939.2 *Direct URL to data*: http://dx.doi.org/10.17632/y258s6d939.2 |

**Value of the Data**

- The dataset includes 12 instances of the three-dimensional Bin Packing Problem (3dBPP, [2]). All the packages that compose each instance have been randomly generated using our own instance generator *Q4RealBPP-DataGen* to avoid any bias. The benchmark is useful for measuring the performance of solvers developed for the same purpose, especially if the solvers rely on a Quantum Processing Unit (QPU).

- Along with the instances, the benchmark also includes a Python script to generate synthetic datasets for the problem. With this generator, researchers can create their own instances for benchmarking purposes.

- Classical Bin Packing related benchmarks are usually composed of large instances, containing few small-sized cases (if any) [3]. *Falkenauer U* or *Schwerin* datasets are well-known examples that confirm this situation, in which smallest instances count with 120 and 100 items. For this reason, researchers working in the quantum computing field can specially benefit from the benchmark proposed in this data article. This is so because of the size of each instance, which is adapted to be solved with current quantum devices.

- Both the instances and the data generator are open source, so they can be modified or extended to other Bin Packing Problem variants [4] with the aim of pushing forward the research in this field.

- The benchmark also includes the results obtained in each instance using the Leap Constrained Quadratic Model Hybrid Solver of D-Wave (`LeapCQMHybrid`, [5]). These results are provided in both image and text format.

**Objective**

The benchmark described in this data article provides 12 instances of real-world oriented 3dBPP scenarios. To properly characterize these realistic industrial use cases, the following requirements have been taken into consideration: *i)* overweight restrictions, *ii)* affinities among package categories, *iii)* preferences in relative positioning and *iv)* load balancing. This is the first quantum-computing oriented benchmark for dealing with the real-world 3dBPP. This is so, because of the sizes of the generated instances, which are adapted to the capacities of current quantum devices. Additionally, there is no benchmark in the literature that addresses all the characteristics covered in this study. In addition to the data provided, and equally important, we present a data generation script, coined *Q4RealBPP-DataGen*, to create new instances.

**Data Description**

The dataset consists of 12 instances for the 3dBPP, each one considering different real-world oriented restrictions. These are the constraints introduced in the benchmark:

- *Item and bin dimensions*: being a three-dimensional problem, packages and bins have an associated length, width, and height, representing dimensions $X$, $Y$ and $Z$, respectively. Items stored in a bin must not exceed its capacity in terms of dimensions, and all the bins in the same instance have the same predefined $[X,Y,Z]$ dimensions.

- *Overweight restrictions*: each item has an associated weight, and bins have a maximum capacity. This restriction requires that the total weight of the stored items assigned to a bin not exceed its maximum capacity.

- *Affinities among package categories*: this restriction introduces positive and negative affinities (incompatibilities) among item categories. This means that items that share a positive affinity must be packed together, while incompatible packages must be assigned to different bins.

- *Preferences in relative positioning*: relative positioning lets the user establish a sorting strategy by package-category location in given axis. For instance, load-bearing must govern the placement of the items with respect to the $Z$-axis. For the sake of simplicity, this could be attained by applying a simple rule: sort the packages based on the mass ratio between packages to decide what item should rest on which one. Anyway, these preferences can accommodate other positioning patterns, such as sorting in $X$-axis according to the delivery schedule.

- *Load balancing*: center of mass to distribute the stored items according to one reference point.

It should be noted that the units of measurement have not been specified as they are not relevant for the study. In search of instances that maximize the difference in performance, each instance has its own particularities, which are summarized in Table 1. Also, Table 2 describes in detail each instance.

*Table 1: Main features of each instance of the benchmark*

| Instance | # of items | Dimensions | Overweight | Pos. Aff | Incom. | Relative Pos (q=6) | L. Balancing |
|---|---|---|---|---|---|---|---|
| 3dBPP_1  | 51 | ✓ |   |   |   |   |   |
| 3dBPP_2  |    | ✓ | ✓ |   |   |   |   |
| 3dBPP_3  | 52 | ✓ |   |   |   |   |   |
| 3dBPP_4  |    | ✓ |   |   |   | ✓ |   |
| 3dBPP_5  | 53 | ✓ |   |   |   |   |   |
| 3dBPP_6  |    | ✓ |   | ✓ |   |   |   |
| 3dBPP_7  | 46 | ✓ |   |   |   |   |   |
| 3dBPP_8  |    | ✓ |   | ✓ | ✓ |   |   |
| 3dBPP_9  | 47 | ✓ |   |   |   |   | ✓ |
| 3dBPP_10 | 51 | ✓ |   |   |   |   | ✓ |
| 3dBPP_11 | 38 | ✓ | ✓ | ✓ | ✓ | ✓ | ✓ |
| 3dBPP_12 | 38 | ✓ | ✓ | ✓ | ✓ | ✓ | ✓ |

*Table 2: Description of the 12 instances that compose the benchmark*

| Instance | Description |
|---|---|
| 3dBPP_1 | 51 items with dimension restrictions. |
| 3dBPP_2 | 51 items with dimension restrictions and a maximum weight capacity of 1000. |
| 3dBPP_3 | 52 packages with dimension restrictions. |
| 3dBPP_4 | 52 items with dimension restrictions. Items with ID {0,1,9} are heavy packages that must be beneath the rest of the items. |
| 3dBPP_5 | 54 packages with dimension restrictions. |
| 3dBPP_6 | 54 items with dimension restrictions. Items with ID {7,9} and {4,7} must not be packed together (mutual incompatibility between first ID and second ID in each set). |
| 3dBPP_7 | 46 items with dimension restrictions. |

| | | |
|---|---|---|
| `3dBPP_8` | 46 packages with dimension restrictions. Items with ID {4,8} must not be packed together, while items with ID {0,3} and {0,8} must be stored in the same bin. |
| `3dBPP_9` | 47 items with dimension restrictions. Center of mass is in the middle of the bin (750, 750). |
| `3dBPP_10` | 47 items with dimension restrictions. Center of mass is in (900, 500). |
| `3dBPP_11` | 38 items with dimension restrictions; maximum weight capacity of 800; items {0,7} are heavy packages; {7,9} incompatible; {0,3} and {0,8} must be packed together; center of mass in (750, 750). |
| `3dBPP_12` | 38 items with dimension restrictions; maximum weight capacity of 900; items {3,4} are heavy packages; {4,8} incompatible; {2,4} must be packed together; center of mass in (500, 500). |

Regarding the format of each instance, for the sake of clarity, we depict in Figure 3 the structure of the `3dBPP_11` instance. This format is an evolution of the one proposed by D-Wave in [6], which in fact served as inspiration for our work. Thus, to build an instance, eight different characteristics should be considered. Table 3 lists these features.

It should be highlighted at this point that, given the previous settings for package definition, and for the sake of simplicity, the constraints are imposed on the item's IDs, which means that the rules described by the constraints apply to all items with the same ID. If users preferred package level assignments, they would have to simply create a dedicated ID for each package.

*Table 3: Features that compose the 3dBPP instances comprised in the benchmark*

| Name | Format | Description | Mandatory |
|---|---|---|---|
| *Max num of bins* | Integer | The maximum number of bins available for storing all the items | ✓ |
| *Bin dimensions* | Array of integers | An array of length 3 to specify bin dimensions: $[X, Y, Z]$ | ✓ |
| *Max weight* | Integer | Maximum capacity of the bins in terms of weight | |
| *Relative Pos* | Dictionary of lists of integer pairs | This value is represented as a dictionary $[q: L]$, in which $q$ stands for the relative positioning that the pairs of integers comprised in list $L$ must follow. As an example, {6: (5,1) (2,1)} means that packages with ID=5 and ID=2 must have the relative position $q = 6$ regarding items with ID=1. In this regard, and for the sake of understandability, $q = 1$ represents "at the left"; $q = 2$ stands for "behind", $q = 3$ is "below", $q = 4$ depicts "at the right", $q = 5$ means "in front", and $q = 6$ represents "above". | |
| *Incompatibilities* | List of integer pairs | Each pair of the list represents an incompatibility, so that $(I,J)$ means that items with ID=$I$ **cannot** be placed in the same bins as items with ID=$J$ | |
| *Positive Affinities* | List of integer pairs | Analogously, each pair of the list represents a positive affinity, meaning that ID=$I$ **must** be placed in the same bins as items with ID=$J$ | |
| *Center of mass* | Pair of integers | This pair of integers are introduced for load balancing purposes, and they represent the *X* and *Y* coordinates in which the items should gravitate. | |
| *Items* | List of items | This list has an entry for each item category available. For each category, six different values should be introduced: the category ID, the number of packages for each category, and the length, width, height, and weight of all the packages in the category. All these values must be integers. | ✓ |

```
# Max num of bins: 2
# Bin dimensions (L * W * H): (900,900,900)
# Max weight: 800
# Relative pos: {6: [(1, 0), (1, 7), (2, 0), (2, 7), (3, 0), (3, 7), (4, 0), (4, 7)
#                   (5, 0), (5, 7), (6, 0), (6, 7), (8, 0), (8, 7), (9, 0), (9, 7)]}
# Incompatibilities: (7,9)
# Positive affinities: (0, 3) (0, 8)
# Center of mass: (750, 750)

  id     quantity     length    width    height    weight
  ----   ---------    -------   ------   -------   -------
   0         6          218      247       216       50
   1         2          215      265        64       20
   2         3          220      296       267       20
   3         6          171      307       101       20
   4         1          280      318       298       20
   5         2          265      321       138       20
   6         3          185      349       157       20
   7         5          297      358       151       50
   8         6          207      362       107       20
   9         4          201      399        96       20
```

*Figure 1: Representation of 3dBPP_11 instance*

The main contribution of the benchmark proposed in this data article is twofold. First, as mentioned before, thanks to the sizes of the generated instances, this is the first quantum-oriented benchmark for solving the 3dBPP. Delving deeper into this aspect, quantum optimization has generated a significant impact in the scientific community. The advances made in the related hardware and the democratization of its access have contributed to the promotion of this scientific area. Anyway, research is restricted by the status of the hardware. There are some limitations on current quantum computers that have a negative impact on their performance. The current state of quantum computing is known as the *noisy intermediate-scale quantum* (NISQ, [7]) era. Quantum devices available in this NISQ era are distinguished by not being fully able to tackle large problems reliably. The evaluation of quantum or hybrid approaches is hampered by this condition, due to the fact that researchers are pushed to build ad-hoc problem instances adapted to the limited capacity of quantum computers. This holds true even when tackling well-known optimization problems, and this circumstance has a direct impact on the capacity to replicate and compare different techniques. More specifically, and focusing on the 3dBPP, the *LeapCQMHybrid* solver of D-Wave, which is one of the most powerful quantum solvers currently available, struggles when dealing with instances composed of more than 75 packages, making the existing datasets not practical for dealing with quantum devices. For this reason, we present in this data article a common-use benchmark for the 3dBPP approachable by the different quantum computers available, and that facilitates the comparison and replicability of the newly proposed methods in the field of quantum optimization.

Secondly, most of the 3dBPP instances that can be openly found in the literature are usually focused on basic variants of the problem, considering just the dimension and weight restrictions[1,2].

---
[1] https://www.euro-online.org/websites/esicup/data-sets/#1535975694118-eedb4714-39e4
[2] https://github.com/Wadaboa/3d-bpp

In this benchmark, affinities among package categories, preferences for package ordering, and load balancing are considered. Also, thanks to the developed *Q4RealBPP-DataGen*, users can generate tailored instances by activating/deactivating constraints suitable for their preferences.

Finally, in addition to the data provided, which can be found in the folder coined as *input*, and the *Q4RealBPP-DataGen* data generator, we provide further material with complementing purposes:

- *Description.txt*: this txt file provides a description of each instance, including the information depicted in Table 2 of this article.
- *Constraints and variables.txt*: this informative file lists how many variables and constraints are needed for correctly modelling each generated instance. These values define the complexity and size of each problem.
- *Output*: with the intention of providing a results baseline, and for the sake of replicability, we provide in this folder the results obtained by a `LeapCQMHybrid` when solving the complete benchmark. The whole experimentation was conducted between February 25 and 27, 2023. For further information about the conducted tests, we refer readers to [8]. For each solved instance, we provide two output files:
  - *name_instance.png*: a graphical representation of the solution provided by the solver.
  - *name_instance_sol.txt*: this file contains descriptive information about the solution provided by the solver. Along with the data of the instance (center of mass, positive affinities, incompatibilities, relative positioning, and number of cases packed), it includes the value of the objective function reached, number of bins used, and the total weight accumulated per bin. Additionally, this file contains the position that each item occupies within the bin. In this regard, for each package, the following information is provided: ID, the bin and the coordinates in which the item is located $(x, y, z)$, the amount of space occupied $(x', y', z')$ and its weight. Figure 2 represents an excerpt of an example output file.

```
# Number of bins used: 2
# Number of cases packed: 38
# Objective value: 2.391
# Max weight: 1000
# Weight of bins: 480.0 391.0
# Relative pos: {6:(1,8)(2,8)(3,8)(4,8)(5,8)(6,8)(7,8)(9,8)}
# Incompatibilities: (0,5) (1,7) (5,6)
# Positive Affinities: (3,4) (4,5)
# Center of mass: (750,750)

id    bin-loc    orientation       x       y     z    x'    y'    z'    weight
----  ---------  -------------  ------  ------  ---  ----  ----  ----  --------
  0       2           4        2131.5   700.5    0   237    99   171      27
  0       2           4        2131.5   601.5    0   237    99   171      27
  0       2           4        2131.5   799.5    0   237    99   171      27
  1       1           2         846     871.5    0   145   159   243      20
  1       1           6         870     628.5    0   159   243   145      20
  1       1           2         870     469.5    0   145   159   243      20
  1       1           1        1029     628.5    0   145   243   159      20
  1       1           6         670.5   846      0   159   243   145      20
  2       2           5        2570     645      0   138   210   265      20
  2       1           5         516     648    120   138   210   265      20
```

*Figure 2: an excerpt of an output file (3dBPP_test_sol)*

## Experimental Design, Materials and Methods

The whole benchmark described in this data article has been built using an ad-hoc Python script (named *Q4RealBPP-DataGen*). Thanks to this script, a user can easily generate additional instances compliant with what is exposed in this article. *Q4RealBPP-DataGen* gives the user the possibility of taking a pre-computed pool of packages (openly available at https://github.com/Wadaboa/3d-bpp) or creating a new set of items from scratch (following the criteria described in [1]). To do so, these parameters have to be set accordingly: $using\_dataset$, $num\_items, min\_width - max\_width$, $min\_length - max\_length$, $min\_height - max\_height$, and $min\_weight - max\_weight$. The rest of the problem is characterised by the following parameters: $num\_bins$, $bins\_dims$, $max - bin\_capacity$, $mass\_ratio, num\_incompatibilities$, $num\_positive\_affinities$, and $CoM$.

Note that incompatibilities and positive affinities are randomly generated, and it is only required for the user to indicate the number of constraints of this nature in *Q4RealBPP-DataGen*. However, this script is conceived to alleviate the creation of a benchmark, and it is not compulsory. Contrarily, a user could opt for manually defining the specific characteristics of the instance directly on a file to enrich the diversity. If the user wants to make use of the load-bearing constraint, he/she should specify the order manually in the dictionary $Relative\ Pos$ for $q = 6$ or take advantage of $mass\_ratio$ parameter shortcut, which will help fill automatically the list in the dictionary.

As a summary, in order to properly use *Q4RealBPP-DataGen*, the parameters described in Table 4 should be considered.

*Table 4: Input parameters for Q4RealBPP-DataGen*

| Parameter | Description |
|---|---|
| num_bins | Number of maximum bins for the instance |
| bins_dims | Bins dimensions (L, W, H) |
| max_bin_capacity | Maximum bin capacity in terms of weight (= $None$ for not considering this feature). |
| mass_ratio | This value, represented as a float greater than 1, is used for load bearing purposes in the following way: if a pair of packages $i$ and $j$ satisfies $weight_i/mass\_ratio > weight_j$, $i$ cannot be placed above $j$. This way, *Q4RealBPP-DataGen* automatically builds the dictionary of values "*Relative Pos*" based on these principles (if the user does not want to contemplate this feature, this value should be = $None$). |
| num_incompatibilities | Number of incompatibilities randomly generated (= 0 for not considering it) |
| num_positive_affinities | Number of positive affinities randomly generated (= 0 for not considering it) |
| CoM | Center of mass $(X, Y)$ |
| using_dataset | A Boolean which indicates if the item categories are extracted from the seed dataset (= $True$) or randomly generated (= $False$). |
| min_width, max_width | If $using\_dataset = False$, the maximum and minimum width for each package category. |
| min_length, max_length | If $using\_dataset = False$, the maximum and minimum length for each package category. |
| min_height, max_height | If $using\_dataset = False$, the maximum and minimum height for each package category. |
| *min_weight, max_weight* | If $using\_dataset = False$, the maximum and minimum weight for each package category. |
| *num_categories* | The number of item categories chosen to participate in the instance, being a product (i.e., category) a description of specific dimensions and weight. |
| *num_items* | The number of items composing the instance. This value helps to create replicates of the products (augmenting the quantity value of each product) |

Finally, with the intention of demonstrating the functionality of *Q4RealBPP-DataGen*, Figure 3 shows an example of the parameterization of the script, and the instance generated after running it. Also, we show in Figure 4 a possible solution for the instances generated for showcasing purposes. We share this instance as well as its solution in the dedicated repository (http://dx.doi.org/10.17632/y258s6d939.1), labeled as `3dBPP_test`.

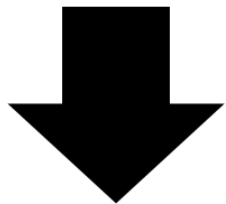

Figure 3: example of an instance generation with Q4RealBPP-DataGen

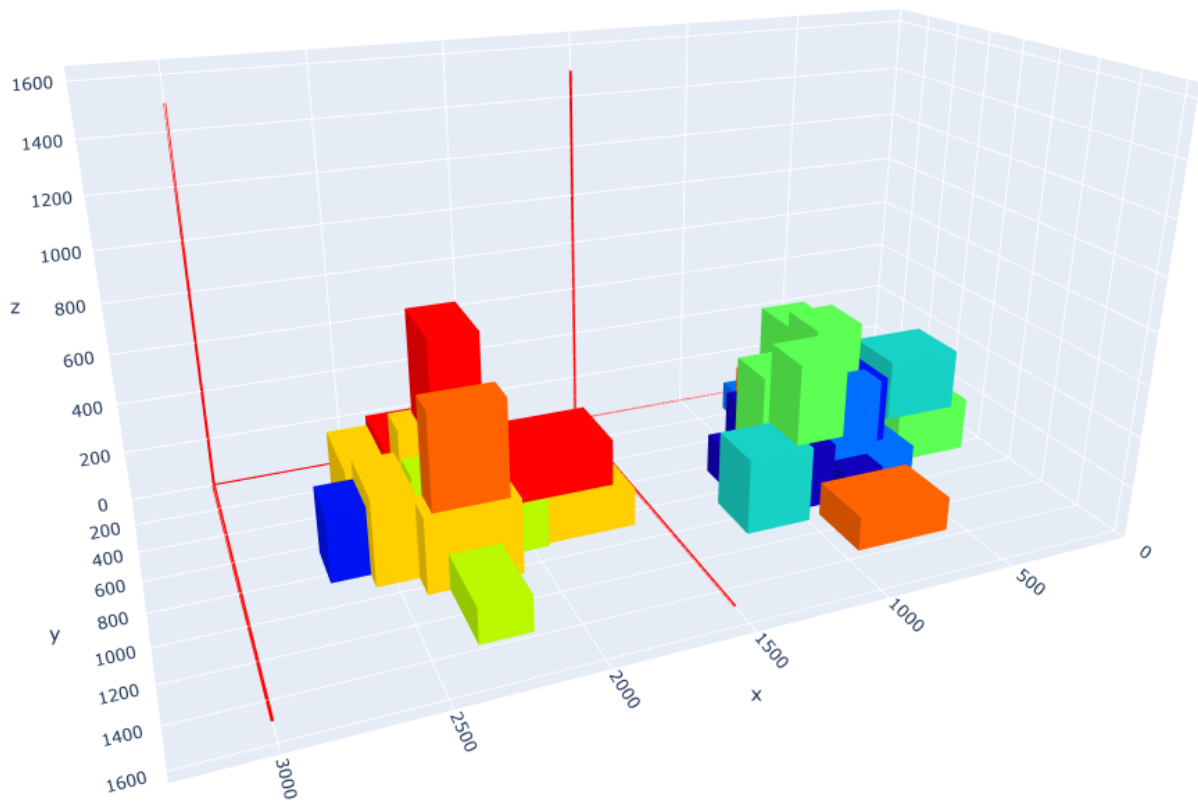

*Figure 4: A possible solution to the instance generated with demonstrating purposes*

## Ethics Statements

This work did not involve studies with animals and humans.

## CRediT author statement

**Eneko Osaba**: Conceptualization, Validation, Writing - Original draft preparation. **Esther Villar-Rodriguez**: Conceptualization, Software preparation, Validation, Writing- Reviewing and Editing. **Sebastián V. Romero**: Conceptualization, Validation, Writing- Reviewing and Editing.

## Acknowledgments

This work was supported by the Basque Government through ELKARTEK program (BRTA-QUANTUM project, KK-2022/00041), and through HAZITEK program (Q4_Real project, ZE-2022/00033). This work was also supported by the Spanish CDTI through Plan Complementario Comunicación Cuántica (EXP. 2022/01341) (A/20220551).

## Declaration of interests

The authors declare that they have no known competing financial interests or personal relationships that could have appeared to influence the work reported in this data article.